\documentclass[runningheads]{llncs}

\usepackage{eccv}

\usepackage{eccvabbrv}

\usepackage{graphicx}
\usepackage{booktabs}
\usepackage{comment}
\usepackage{csquotes}
\usepackage{multirow}
\usepackage{subcaption}

\usepackage[accsupp]{axessibility}  

\definecolor{ourgreen}{HTML}{d2f0aa}
\definecolor{ouryellow}{HTML}{fcef9e}

\usepackage[colorinlistoftodos,prependcaption,textsize=tiny]{todonotes} 
\usepackage{hyperref}

\usepackage{orcidlink}

\begin{document}

\title{Audio-driven Talking Face Generation with Stabilized Synchronization Loss} 


\author{Dogucan Yaman\inst{1}\orcidlink{0000-0002-5047-295X} \and
Fevziye Irem Eyiokur\inst{1}\orcidlink{0000-0001-5754-5405} \and
Leonard Bärmann\inst{1}\orcidlink{0000-0003-3092-8726} \and \\
Haz{\i}m Kemal Ekenel\inst{2} \and
Alexander Waibel\inst{1,3}}

\authorrunning{D. Yaman et al.}

\institute{Karlsruhe Institute of Technology, Karlsruhe, Germany \and
Istanbul Technical University, Istanbul, Turkey \and
Carnegie Mellon University, Pittsburg PA, USA\\
\email{dogucan.yaman@kit.edu}}

\maketitle

\begin{abstract}
  Talking face generation aims to create realistic videos with accurate lip synchronization and high visual quality, using given audio and reference video while preserving identity and visual characteristics. In this paper, we start by identifying several issues with existing synchronization learning methods. These involve unstable training, lip synchronization, and visual quality issues caused by lip-sync loss, SyncNet, and lip leaking from the identity reference. To address these issues, we first tackle the lip leaking problem by introducing a silent-lip generator, which changes the lips of the identity reference to alleviate leakage. We then introduce stabilized synchronization loss and AVSyncNet to overcome problems caused by lip-sync loss and SyncNet. Experiments show that our model outperforms state-of-the-art methods in both visual quality and lip synchronization. Comprehensive ablation studies further validate our individual contributions and their cohesive effects. 
  
\end{abstract}

\newcommand{\appRef}[1]{App. \textcolor{red}{#1}}

\section{Introduction}
\label{sec:intro}

Audio-driven talking face generation aims at generating a video with respect to given face and audio sequences.
The objective is to achieve synchronized lip movements corresponding to the provided audio while preserving the identity and visual details.
This task has recently attracted significant attention due to its versatile applications, including dubbing in the film industry, online education, enhancing video conferencing, and dubbing for various types of videos~\cite{zhan2021multimodal,zhen2023human}.

The talking face generation task comprises two primary aspects: (1)~lip synchronization and (2)~visual quality of the face.
Since lips that are out-of-sync with audio can be easily identified by humans, synchronized lips are key to achieving natural and realistic talking-face generation. 
For this, the primary solution is to evaluate audio-lip synchronization and use it as a training objective. 
Wav2Lip~\cite{prajwal2020lip} introduced an improved version of SyncNet~\cite{chung2017out}, a pretrained model designed to measure audio-visual synchronization. 
This model is also used during training to extract features and compute lip-sync loss. 
Subsequently, many approaches have employed improved SyncNet~\cite{prajwal2020lip} to guide the model throughout training.
The main concept is to provide audio-video pairs to SyncNet, 
which extracts features through its image and audio encoders.
Subsequently, cosine similarity with binary cross-entropy loss is measured between these two features~\cite{prajwal2020lip}.
Since the model was trained in this manner, a higher cosine similarity is expected when the lips are synchronized with the given audio.
Although existing approaches with this strategy  
mostly surpass other methods in lip synchronization,  
there are still challenges that must be addressed to enhance performance. 

\begin{figure}[tb]
  \centering
  \begin{subfigure}{0.34\linewidth}
    \includegraphics[width=\linewidth]{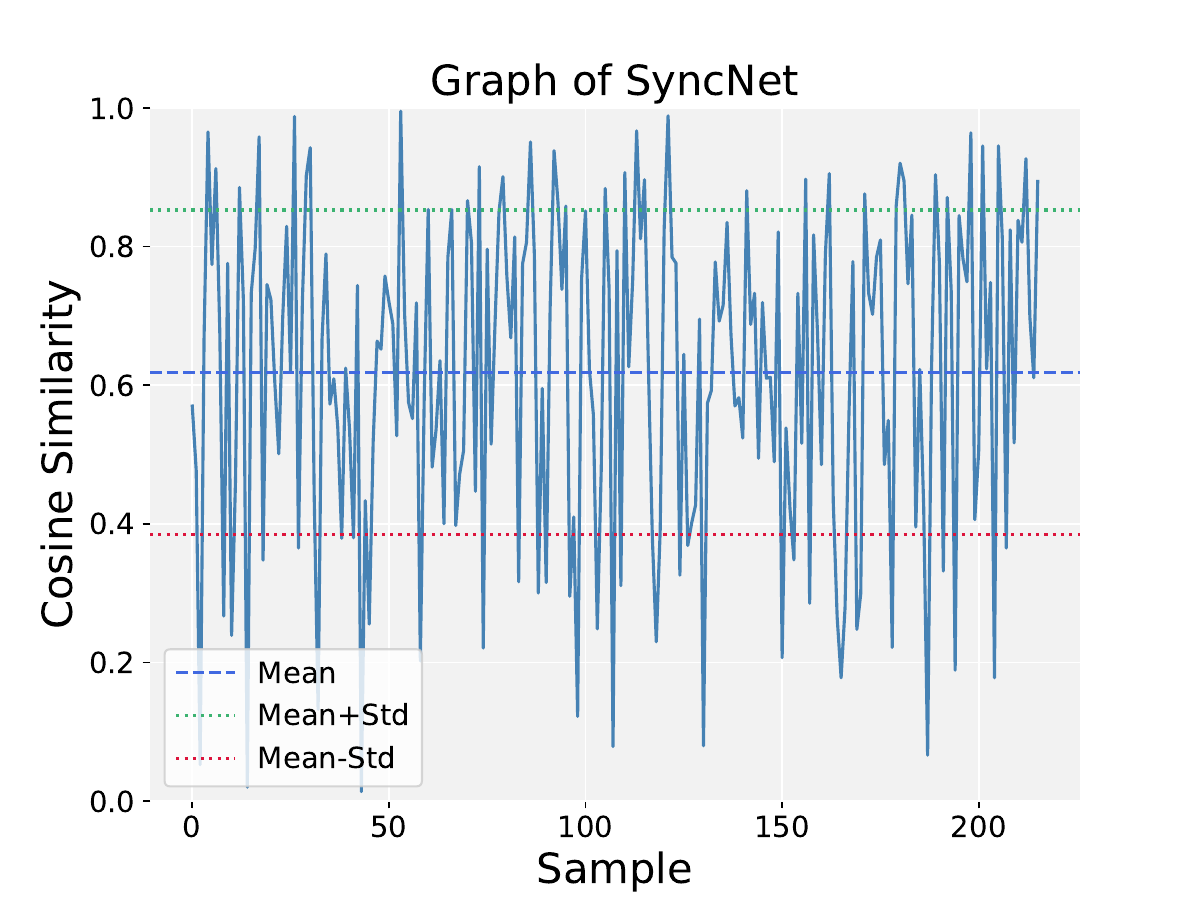}
    \caption{SyncNet~\cite{prajwal2020lip}}
    \label{fig:cosine_syncnet}
  \end{subfigure}
  \begin{subfigure}{0.34\linewidth}
    \includegraphics[trim={0cm 0cm 0cm 0cm},clip,width=\linewidth]{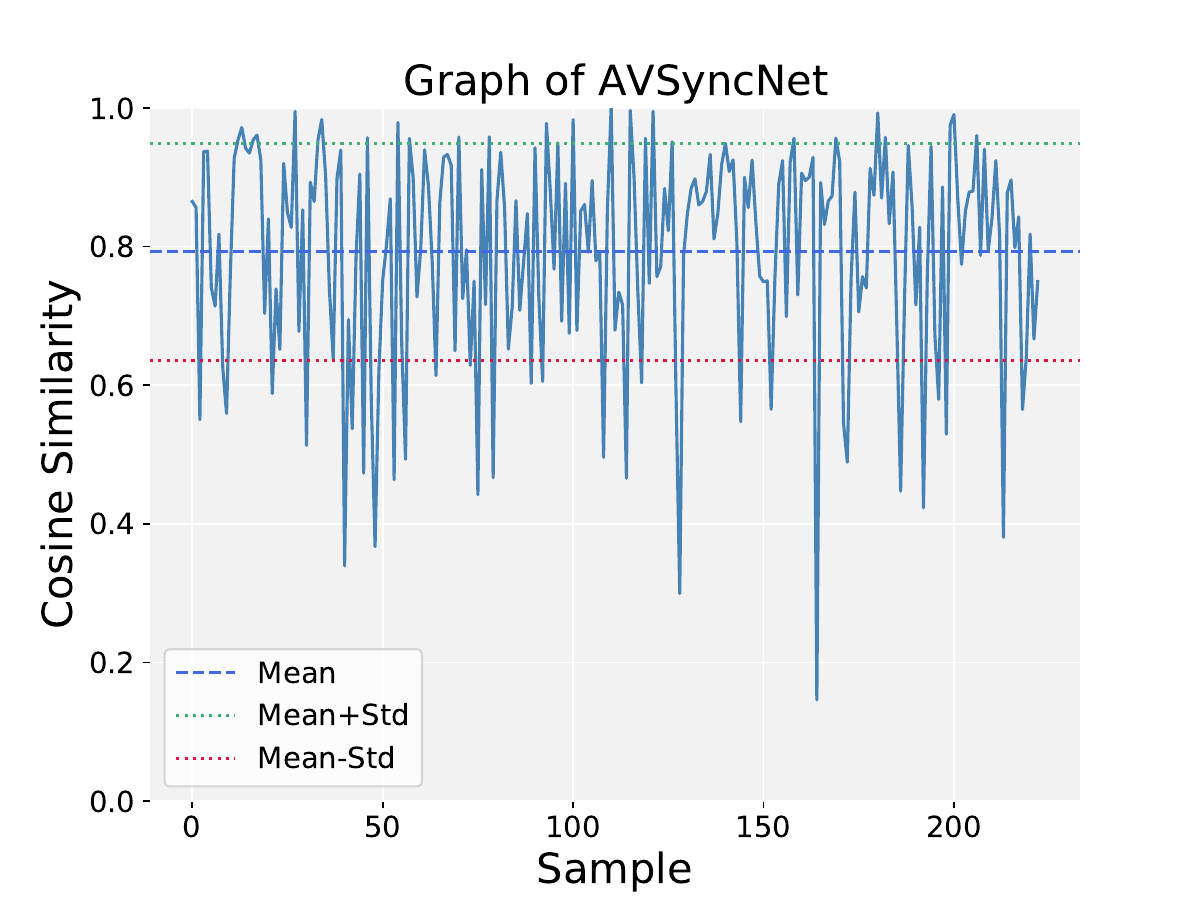}
    \caption{AVSyncNet (ours)}
    \label{fig:cosine_avsyncnet}
  \end{subfigure}
    \begin{subfigure}{0.28\linewidth}
    \includegraphics[trim={0cm 0cm 0cm 0cm},clip,width=\linewidth]{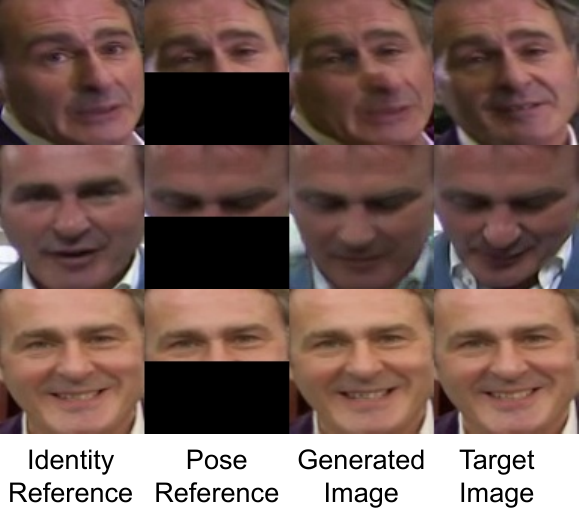}
    \caption{Leaking problems}
    \label{fig:problems}
  \end{subfigure}
  
  \caption{\textbf{(a, b)} Cosine similarity between GT audio-lip pairs on random LRS2 samples, showcasing the instability of SyncNet and more robust performance of AVSyncNet. \textbf{(c)} illustrates full mouth region / lip leaking from the reference, pose effect from the reference, and similar identity reference-target image scenarios.}
  \label{fig:problem_analysis}
\end{figure}

In this work, we identify two main challenges in existing approaches that restrict models from achieving satisfactory performance in both lip synchronization and visual quality: \emph{SyncNet instabilities} and \emph{lip leaking}.
First, in alignment with previous research \cite{nayak2022deep}, we observe instabilities in the performance of SyncNet \cite{prajwal2020lip} on pairs of ground-truth (GT) lip and audio (see \cref{fig:cosine_syncnet} and \appRef{A}).
Thus, when SyncNet is utilized as part of talking face generation training, it might provide an inadequate training signal, assigning low similarity scores to generated images even when the lips are synchronized.
This problem causes unstable training, degrading the lip generation capability of the network, thereby ending up in out-of-sync lips or suboptimal lip-sync. 
Furthermore, the lip-sync loss \cite{prajwal2020lip} and reconstruction losses are conflicting \cite{muaz2023sidgan}.
This leads models to have either poor lip synchronization or degraded visual quality (sometimes even both), escalating further when lip-sync loss and SyncNet are employed with high-resolution (HR) data \cite{zhang2023dinet,muaz2023sidgan}.
To address these problems, we first improve SyncNet and introduce \emph{AVSyncNet}, demonstrating a more robust performance (see \cref{fig:cosine_avsyncnet}) and also overcoming poor shift-invariance characteristics of SyncNet (see \cref{fig:ablation_shifting}). However, despite improved performance, the instability problem is not fully solved (see \cref{fig:cosine_avsyncnet}).
To overcome this problem further, we also introduce a \emph{stabilized synchronization loss}. 
Specifically, instead of directly using the similarity of the (generated lips, audio) pair, we calculate the difference of the similarities between (GT lips, audio) and (generated lips, audio).
We hypothesize that this alleviates the described problems since it guides the model to generate a lip movement with a similar synchronization score as for the GT face.
Together with AVSyncNet, this method empirically enhances the lip synchronization performance as well as avoids visual quality issues caused by misleading lip-sync loss.

Second, we address another main problem in the talking face generation literature: \emph{lip leaking}.
The current gold standard in 2D-based methods is to input a bottom-half masked face image (\enquote{pose reference}), wherein the model is expected to generate the input face with proper lip movements.
However, because of masking, the model requires a reference image to retain the identity and texture in the masked part.
This is achieved by randomly choosing a face image from a different part of the video sequence, referred to as \enquote{identity reference}.
However, this introduces a new challenge:
As it is randomly selected, the lip movements of the identity reference can frequently be similar to the GT lips during training (see last row of \cref{fig:problems}).
Hence, for faster convergence, the model tends to replicate the lip movements from the identity reference, resulting in poor lip synchronization or complete out-of-sync output.
By following the literature \cite{prajwal2020lip,nayak2022deep}, we term this phenomenon as \textit{lip leaking}.
To tackle this problem, we propose a simple yet effective technique: \textit{silent-lip generator}.
The concept involves modifying the identity reference to make the lips closed (thus \enquote{silent-lip}) before feeding it to the talking face generator network.
This method ensures consistently closed lips in the identity reference, effectively mitigating the lip leaking problem.
Our contributions are as follows: (i) We identify and analyze various fundamental problems that harm lip synchronization learning and also cause visual quality issues. (ii) We present a robust and shift-invariant AVSyncNet, and stabilized synchronization loss to overcome the problems caused by lip-sync loss and SyncNet. (iii) We present a silent-lip generator to generate an identity reference with closed lips before feeding the talking face generator to alleviate the lip leaking.

\section{Related Work}
\label{sec:relatedwork}

\textbf{Audio-driven Talking Face Generation.}
Initial studies mapped audio features to time-aligned facial motions~\cite{yehia1998quantitative} or predicted facial motions by HMM~\cite{brand1999voice}.
In~\cite{suwajanakorn2017synthesizing}, videos were generated by finding the images most aligned with the audio. 
ATVGNet~\cite{chen2019hierarchical} transfers audio to facial landmarks and uses pixel-wise loss with an attention mechanism to avoid the jittering problem and leaking of irrelevant speech. 
\cite{das2020speech} and \cite{zhou2020makelttalk} use facial landmark representation to synthesize faces with synchronized lips.
Recent papers address the task as conditional inpainting by masking the bottom half of the input face and feeding the network an identity reference from another time step of the same video, along with the audio segment. 
Wav2Lip~\cite{prajwal2020lip} proposes SyncNet and lip-sync loss to predict lip synchronization and achieves significant performance. 
PC-AVS~\cite{zhou2021pose} introduces a pose-controllable audio-visual system, while GC-AVT~\cite{liang2022expressive}, EAMM~\cite{ji2022eamm} and  EVP~\cite{ji2021audio} control the emotion by utilizing an emotion embedding. 
On the other hand, SyncTalkFace~\cite{park2022synctalkface} uses Audio-Lip Memory to store lip motion features and retrieves them as visual hints for better synchronization. 
VideoReTalking~\cite{cheng2022videoretalking} proposes to manipulate the reference image to have a face with canonical expression to alleviate the sensitivity of the model against the identity reference image. 
Similarly, we introduce a silent-lip generator to implicitly learn to manipulate the lips of the identity reference to mitigate the lip leaking problem.
Compared to the method proposed in \cite{cheng2022videoretalking}, our method works more efficiently to preserve the identity and visual quality.
More recently, LipFormer~\cite{wang2023lipformer} uses a pre-learned facial codebook to generate HR videos, while DINet~\cite{zhang2023dinet} proposes to use a deformation module to obtain deformed features to enhance lip synchronization and head pose alignment.
IPLAP~\cite{zhong2023identity} shows satisfactory visual quality and stability by using intermediate landmark representation and motion field. 
Recently, TalkLip~\cite{wang2023seeing} introduces a global audio encoder, trained with self-supervised learning, to encode features by considering the entire content of the audio. 
Besides, they propose to use lip reading during the training as well as in the evaluation to control whether the content is preserved. 
SIDGAN~\cite{muaz2023sidgan} performs important analyzes, and introduces shift-invariant APS-SyncNet and training objectives along with the coarse-to-fine pyramid model for HR dubbed video generation. 
Recent works use diffusion model because of its stability and accuracy~\cite{shen2023difftalk,stypulkowski2024diffused}.
Finally, in~\cite{waibel2023face}, talking face generation is used as a part of a full system to perform end-to-end face dubbing, involving speech recognition, translation, speech, and video generation.
In contrast to the above, 3D-based and Neural Radiance Fields-based (NeRFs) methods typically generate the entire head, rather than just manipulating 2D images of the face, often involving manipulation of pose, emotion, and 3D face model \cite{blanz1999morphable,booth2018large,zhou2019talking,zhou2020makelttalk,thies2020neural,wu2021imitating,zhang2021facial,zhang2021flow,yin2022styleheat,guo2021ad,yao2022dfa,liu2022semantic,shen2022learning,tang2022real,song2022everybody,papantoniou2022neural,ye2023geneface,wang2023progressive}.
Similarly, portrait animation aims at utilizing a single input image to generate a video by predicting the pose and expression of the subsequent frames, along with ensuring synchronized lips.
Nevertheless, these tasks significantly differ from face dubbing in terms of methodology, task definition, goals, and real-world applications.

\subsubsection{Lip Synchronization.} 
Some earlier works~\cite{hershey1999audio,slaney2000facesync} used hand-crafted features and statistical models to evaluate lip synchronization. 
Recent studies 
proposed to use mutual information between audio-visual features to produce \textit{sync} or \textit{out-of-sync} output for sound~\cite{owens2018audio, chen2021audio,iashin2022sparse} or speech~\cite{chung2017lip, chung2019perfect, afouras2020self, kim2021end, kadandale2022vocalist}. 
While some methods learn lip synchronization implicitly~\cite{suwajanakorn2017synthesizing,kr2019towards,jamaludin2019you,guo2021ad,wu2021imitating,zhang2021flow,chen2019hierarchical}, other methods employ distance between landmarks or facial parameters~\cite{zhou2020makelttalk,papantoniou2022neural,ji2021audio,song2022everybody}. 
Contrarily, the majority of works~\cite{song2018talking,zhou2019talking,zhou2021pose,yao2022dfa,vougioukas2020realistic,eskimez2020end,prajwal2020lip,park2022synctalkface,sun2022masked,zhang2023dinet,liang2022expressive,wang2023lipformer,guan2023stylesync,wang2023progressive,wang2023seeing} extract audio-visual features with an additional network (mostly SyncNet~\cite{prajwal2020lip}) for audio-lip synchronization prediction. Afterwards, while some of these methods utilize lip-sync loss~\cite{prajwal2020lip}, others benefit from contrastive learning by using infoNCE~\cite{oord2018representation}.
We follow a similar strategy by utilizing an additional network to calculate a synchronization loss.
We propose a robust and shift-invariant AVSyncNet along with a stabilized synchronization loss to overcome existing challenges. 

\newcommand{\rulesep}{\unskip\ \vrule\ }
\begin{figure}[tb]
  \centering
  \begin{subfigure}{0.78\linewidth}
    \includegraphics[width=\linewidth]{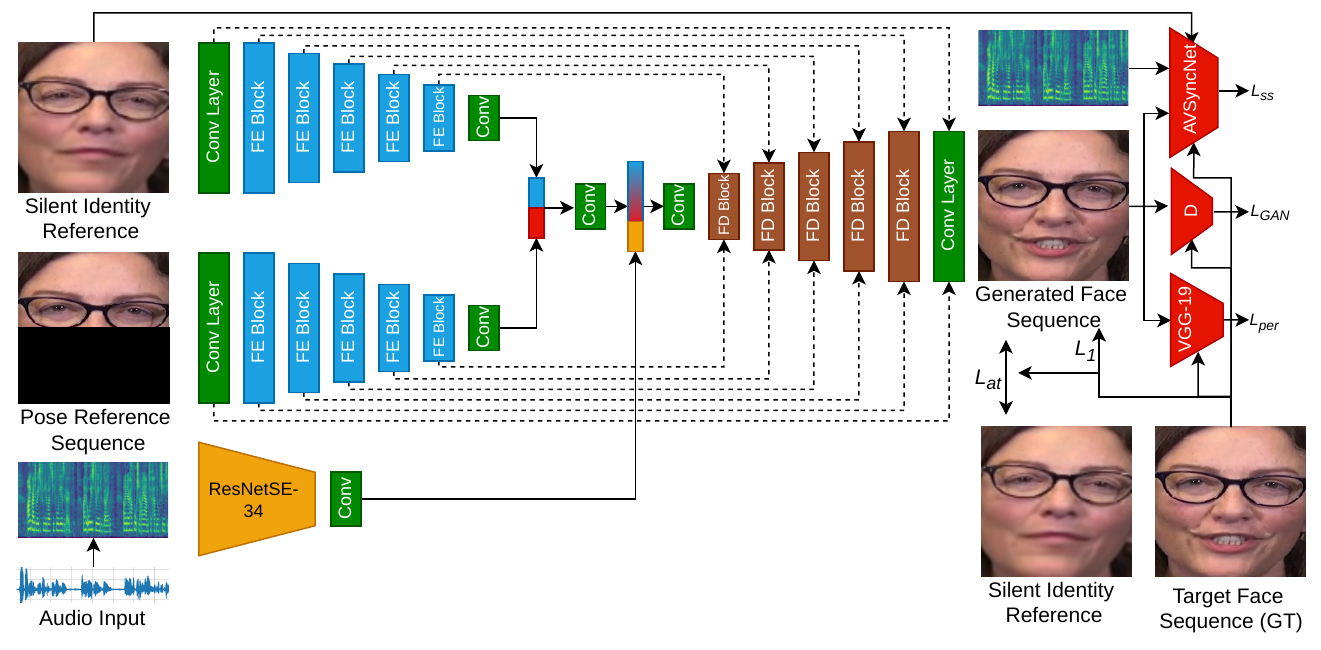}
    \caption{Talking face generation model}
    \label{fig:method_a}
  \end{subfigure}
  \hfill
  \rulesep
  \begin{subfigure}{0.19\linewidth}
    \includegraphics[trim={2cm -4cm 2cm 0cm},clip,width=\linewidth]{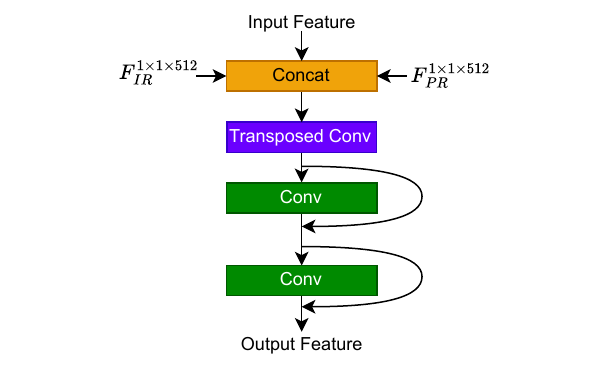}
    \caption{FD block}
    \label{fig:method_b}
  \end{subfigure}

  \caption{Talking face generation model $G_L$ (a) and face-decoding (FD) block (b). Our model receives a pose reference sequence, mel-spectrogram of an audio snippet, and a silent identity reference, that is generated by our silent-lip generator $G_S$, aiming to alleviate lip leaking problem. The model then synthesizes the talking face sequence to ensure lip synchronization. Subsequently, the employed loss functions are computed.}
  \label{fig:short}
\end{figure}

\section{Proposed Approach}
\label{sec:approach}

\subsection{Talking Face Generation}

We propose an audio-driven talking face generation model $ G_L $ with enhanced lip synchronization.
As shown in \cref{fig:method_a}, our model incorporates:~1)~an audio encoder for processing the audio snippet,~2)~an identity encoder for addressing an identity reference image, and~3)~a pose encoder for utilizing a pose reference.

\subsubsection{Audio Encoder.}
The audio encoder $E_A$ generates phoneme-level embeddings, serving as conditions for the face generator to generate lip movements accurately.
Our audio encoder extracts embeddings $ F_A = E_A(A) \in \mathbb{R}^{1 \times 1 \times 512} $ from the given mel-spectrogram $A$, representing the driving audio.
In contrast to existing methods \cite{prajwal2020lip,cheng2022videoretalking,zhang2023dinet,zhong2023identity}, we propose using a pretrained, frozen audio encoder, concurrently trained with a face encoder to learn lip synchronization similar to the objective in SyncNet \cite{prajwal2020lip}.
Thus, we can obtain improved embeddings during the talking face generation training by leveraging the capacity of the pretrained robust audio encoder. We obtain the best score with such frozen audio encoder.

\subsubsection{Face Encoder.}
In accordance with the gold standard in the literature, we utilize an identity reference, $ I^R $, and a pose reference, $ I $, as inputs to the model.
The identity reference is a face image of the subject that provides identity information.
It is different from the pose reference and randomly selected from the same video. 
The pose reference is identical to the target image, except for the bottom half, which is masked, as the model is designed to focus on generating lip movements.
Unlike most conventional methods that employ a joint encoder for processing identity and pose references, 
we use individual encoders to allow each encoder to focus solely on their respective tasks~\cite{muaz2023sidgan}.
Therefore, we utilize two parallel CNN-based face encoders to process identity and pose references individually.
This approach yields better feature representation and ultimately leads to improved performance.

\begin{figure}[tb]
  \centering
  \begin{subfigure}{0.45\linewidth}
    \includegraphics[width=\linewidth]{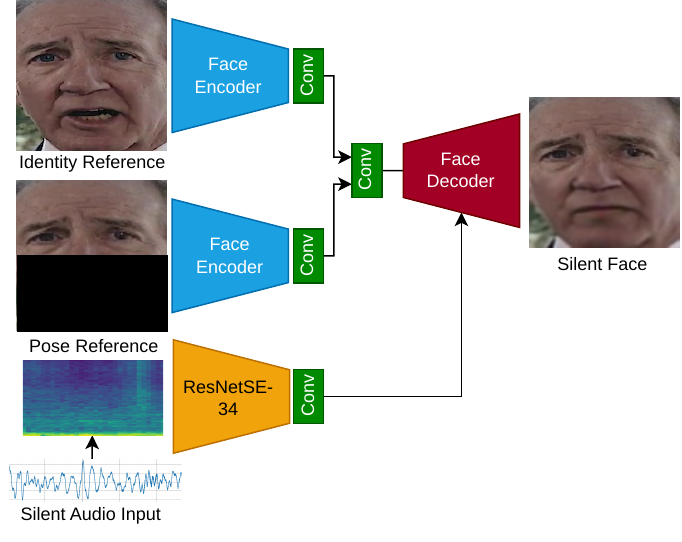}
    \caption{Silent-lip generation model $ G_S $}
    \label{fig:silent_lip}
  \end{subfigure}
  \hfill \hfill 
  \begin{subfigure}{0.4\linewidth}
    \includegraphics[trim={0 -2cm 0 0cm},clip,width=\linewidth]{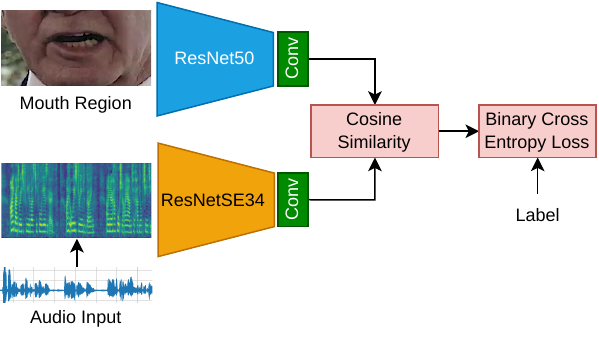}
    \caption{AVSyncNet }
    \label{fig:avsyncnet}
  \end{subfigure}
  \caption{$ G_S $ in inference (a) and AVSyncNet training pipeline (b).}
  \label{fig:additional_models}
\end{figure}

\subsection{Silent-Lip Generator}

Talking face generation involves using an identity reference to preserve the identity in the generated image.
This is particularly important because the bottom half of the pose reference, specifically the mouth region, is masked.
This masking is necessary as we aim to synthesize appropriate lip movements corresponding to the provided audio.
However, models are unintentionally affected by the lip movement of the identity reference, rather than solely gathering identity information (see \cref{fig:problems} and \appRef{C} for details).
This behavior, which we refer to as \textit{lip leaking}, leads to poor lip synchronization or occasionally even non-converged training.
We hypothesize that there are two main reasons for this:
First, the lip movement of the identity reference may occasionally resemble that of the lips in the target image (see last row of \cref{fig:problems}). Hence, the model can lower the synchronization loss more quickly by undesirably replicating the lip movements of the identity reference.
Second, the diversity of lip movements in the identity reference may yield the model to seek a correlation with the target lips.
This causes a challenging disentanglement task for the identity encoder ---namely distinguishing identity information and lip movement.

To mitigate the aforementioned issue, we propose to use an additional model, called silent-lip generator $ G_S $, prior to the talking face generation model $G_L$, aimed at modifying the lip shape of the identity reference. 
Specifically, we reconstruct the input face with closed, flat lips.
This strategy reduces the likelihood of having lips similar to the target face in the identity reference and resolves the issue of diverse lip movements.
Consequently, the model no longer replicates the lips from the identity reference, resulting in stable training \& improved lip sync.

To implement $G_S$, 
we structure the task and the model similar to the talking face generation.
Specifically, we input a bottom-half masked pose reference, an identity reference, and an audio snippet to train a GAN~\cite{goodfellow2014generative} for generating a talking face.
The model is trained to reconstruct the face using both the pose and identity references. 
Notably, we exclude any synchronization loss and SyncNet during training and train the model under the weak condition.
Consequently, the model focuses solely on generating lip movements without synchronization when speech is present.
Therefore, it implicitly learns to generate closed lips when the input audio is silent as shown in \cref{fig:silent_lip}.
Furthermore, by eliminating synchronization loss, we overcome unstable training and the issue of lip leaking from the identity reference for this network.
We choose this approach for its efficient utilization of the same model and data.
Given the scarcity of closed-lip faces in the dataset, we avoid using these frames directly to maintain effective training and generalization.
Please note that we use the same architecture for $G_S$ and $G_L$. 
We initially train $ G_S $ separately on the same training data (LRS2) and then incorporate it into the training of $ G_L $ without further finetuning.
Specifically, we then only pass silent audio to $G_S$, so that it modifies given identity references to have silent lips for subsequent use as identity reference for $G_L$.

\subsection{Video Generation}
For the talking face video generation, we employ the aforementioned components.
Initially, we employ our pretrained silent-lip generator $ G_S $ to synthesize the identity reference with closed lips.
Subsequently, we input this silent identity reference to the identity encoder and the pose reference to the pose encoder in the talking face generation model $ G_L $.
Similarly, we provide the mel-spectrogram of the corresponding audio snippet to the audio encoder.
Next, we concatenate the embeddings from the identity and pose encoders, along with the depth dimension, and pass it through a $ 1 \times 1 $ convolution layer to reduce the depth.
Finally, we concatenate this feature representation with the audio embeddings before feeding them into the face generator.
The face generator generates the entire face with accurate lip movements by preserving the identity and the pose.

We illustrate the talking face generation model $ G_L $ in \cref{fig:method_a}.
We integrate the U-Net architecture \cite{ronneberger2015u} for our overall design, leveraging its adequate performance in reconstruction tasks while ensuring computational efficiency.
Our identity encoder and pose encoder share the same architecture, consisting of consecutive face-encoding (FE) blocks.
Each block has a strided-convolutional layer followed by two non-strided convolutional layers, each paired with a batch normalization layer \cite{ioffe2015batch} and a ReLU activation function \cite{nair2010rectified,krizhevsky2012imagenet}.
We also use the residual connection strategy \cite{he2016deep} by summing the input and output of each block before forwarding it to the next layer.
On the other hand, our face generator has consecutive face-decoding (FD) blocks.
As shown in \cref{fig:method_b}, within each block, we utilize a transposed-convolutional layer, followed by two convolutional layers incorporating batch normalization and ReLU activation function.
Moreover, we apply a skip connection between the reciprocal layers of the face encoders and the decoder to retain high-level features and enhance the training stability.

\subsubsection{GAN Loss.} 
To train our model, we utilize GAN loss \cite{goodfellow2014generative} and employ a discriminator, which is a straightforward CNN-based binary classification network to distinguish real and fake samples, designed with a balanced architecture aligned with our face encoders.
We benefit from consecutive strided-convolutional layers followed by the Leaky ReLu activation function and spectral normalization \cite{miyato2018spectral}. 

\subsubsection{Reconstruction Loss.} 
We employ L1 loss in pixel space $ L_{pixel} = ||I' - I^{GT}||_1 $ between the generated $I'$ and the target faces $I^{GT}$  to ensure consistency in areas outside the lips and maintain the illumination condition.
We further utilize perceptual loss \cite{johnson2016perceptual} based on the pretrained VGG-19 \cite{simonyan2014very}:
\newcommand{\VGG}[0]{\textsc{VGG}}
\begin{equation}
    L_{per} = \sum_{i=1}^{5} c_i ||\VGG^{\phi_i}(I') - \VGG^{\phi_i}(I^{GT})||_1
    \label{eq:L_per}
\end{equation}
where $ c_i $ are weight coefficients from \cite{johnson2016perceptual}, and $ \phi $ indicates the set of 
VGG layers.

\subsubsection{Adaptive Triplet Loss.}
Although our goal is to capture visual details from the identity reference, we observe that the model occasionally focuses on the visual details (e.g., illumination, pose) in the identity reference excessively and this tendency could potentially degrade the quality and stability of the generated face sequence, resulting in suboptimal performance.
To tackle this, we exploit a triplet loss strategy \cite{schroff2015facenet}, aiming to minimize the distance between the generated face and GT, while maximizing the distance between the generated face and the identity reference.
However, the random selection of identity references increases the probability of choosing an image that closely resembles the GT.
This scenario poses a challenge for the vanilla triplet loss, potentially degrading training and resulting in poor visual and pose quality.
To mitigate this, we introduce an adaptive triplet loss that considers the similarity between the identity reference and GT during loss computation to alter its effect.
The formula is as follows:
\begin{equation}
    L_{at} = \left [ D(\VGG(I'), \VGG(I^{GT})) - \frac{D(\VGG(I'), \VGG(I^R))}{D(\VGG(I^{GT}), \VGG(I^R))} + \alpha \right ]_+
\end{equation}
where $ [\cdot]_+ = max(\cdot, \epsilon) $, $ D $ represents the L2 distance, and we empirically choose $ \alpha = 1 $.
In this loss, we leverage the ratio of the similarity between the generated image and identity reference to that of the GT and the identity reference to adjust the loss value.
As the identity reference becomes more similar to GT, the impact of the distance between the generated image and identity reference on the loss diminishes, since expecting a high distance in this case is not reasonable.
Since our objective is to incorporate visual details from the identity reference, we opt for a very low coefficient to avoid conflicting with the primary goal.

\subsection{Learning Synchronization}
\label{avsyncnet}
The lip-sync loss \cite{prajwal2020lip} serves as a method to calculate synchronization between audio and video.
Leveraging the pretrained SyncNet \cite{prajwal2020lip} for feature extraction from both audio and video inputs 
demonstrates reasonable performance in learning lip synchronization.
However, the SyncNet is significantly unstable when measuring this similarity.
Our evaluation using SyncNet on GT training data reveals notable fluctuations in cosine similarity between video and audio, contrary to the expected high scores (see \cref{fig:cosine_syncnet} and \appRef{A} for details).
Therefore, this provides conflicting information to the system, resulting in poor lip synchronization, unstable training, and degraded visual quality.
To tackle this problem, we present a more accurate, shift-invariant, and robust version of SyncNet, named AVSyncNet, and introduce a novel, stabilized synchronization loss.

\subsubsection{AVSyncNet.}
We employ a ResNet-50-based \cite{he2016deep} image encoder, known for its superior performance in face recognition, alongside a ResNetSE-34-based audio encoder \cite{chung2020defence}, which is a modified version of ResNet-34 \cite{he2016deep} designed to handle spectrogram inputs.
This design makes the AVSyncNet model robust and shift-invariant compared to SyncNet~\cite{prajwal2020lip}.
We train our model on the LRS2 training data \cite{LRS2}, calculating the cosine similarity between audio and lip features, followed by a binary cross-entropy loss, as shown in \cref{fig:avsyncnet}.
During each training step, we provide a set of images (5 images) along with the corresponding audio.
For negative samples, we randomly select an audio snippet from the non-overlapping part of the video.
Please note that as we feed the bottom half of the face to the image encoder, we adapt the first layer of ResNet-50 for an input size of $ 112 \times 224 $.

\subsubsection{Stabilized Synchronization Loss ($ L_{ss} $).}
Although AVSyncNet shows improved performance compared to SyncNet~\cite{prajwal2020lip} and alleviates existing instability problem that harms lip-sync and visual quality, 
the unstable performance is not fully solved due to the inherent challenges of the task (see \cref{fig:cosine_avsyncnet} and \appRef{A, B}).
Therefore, we introduce a stabilized synchronization loss ($ L_{ss} $) to improve the lip synchronization performance further in conjunction with AVSyncNet by providing more stable and precise supervision.
The formula is shown below: 
\newcommand{\cossim}[0]{\text{cossim}}
\newcommand{\avsim}[0]{\textsc{AVsim}}
\begin{equation}  
    L_{ss} = -\log \left ( 1 - \frac{|x-y| + \epsilon}{|x-y| + |y-d| + \epsilon} \right )
\end{equation}
\begin{align}
    x = \avsim\left(I', \;  A\right), \quad
    y = \avsim\left(I^{GT}, \;  A\right), \quad
    d = \avsim\left(I^R, \;  A\right)
\end{align}
where $ I' $, $ I^{GT} $, and $I^R$ are generated, GT, and identity reference lips, respectively, while $ A $ is their corresponding audio.
$\avsim(I, A)$ indicates the audio-visual similarity between a face image (bottom half only, \ie lips) and audio, given by the cosine similarity of extracted image and audio features $ \phi_{AVS}^V(I) $, $ \phi_{AVS}^A(A) $ of the respective AVSyncNet encoder.

In this formulation, $ x $ followed by cross-entropy loss denotes the lip-sync loss \cite{prajwal2020lip}.
However, to address the unstable and fluctuating performance, we utilize the relative distance in similarity between GT lips-audio and generated-audio pairs \footnote{This is reminiscent of distillation loss~\cite{hinton2015distilling}, as the actual value of the scores is neglected, and only their difference provides loss value for the training. However, initial experiments trying to directly minimize the distance between SyncNet (or AVSyncNet) image encoder features of generated and GT faces showed poor performance. We hypothesize that this is caused by SyncNet image features being only meaningful for comparing with corresponding audio features due to its training strategy.}. 
Sometimes, randomly selected reference images may have a lip movement similar to that of the target image.
We have already introduced the silent-lip generator $ G_S $ to mitigate this situation.
However, AVSyncNet, while less severe than SyncNet, might still be unstable (\eg, unexpectedly high scores with incorrect pairs or vice versa), resulting in a minor lip leaking and stability issue. 
This problem arises when AVSyncNet assigns an erroneously high score to silent lip-audio pairs as well as when the target lip shape looks similar to closed lips.
To mitigate this, we inject the similarity score between the identity reference and audio into the formulation.
Specifically, we penalize the model more when the identity reference-audio pair shows higher similarity.

\subsection{Implementation Details}

Combining all the presented contributions, the total loss is:
\begin{equation}
    L = L_{GAN}(G, D) + \lambda_1 L_{pixel}(G) + \lambda_2 L_{per}(G) + \lambda_3 L_{ss}(G) + \lambda_4 L_{at}(G)
\end{equation}
where $ G $ and $ D $ indicate generator and discriminator outputs, respectively.
We empirically found the best coefficients as $ (\lambda_1, ..., \lambda_4) = (10, 1, 2, 0.5) $.
Please note that this is the loss function for training $G_L$, while for $G_S$ we set $ \lambda_3 = \lambda_4 = 0 $.

We process videos by using $ 5 $ consecutive frames in each step to consider the temporal information.
We detect faces with FAN \cite{bulat2017far}, followed by acquiring tight crops and resizing to $ 96 \times 96 $, as faces in LRS2 \cite{LRS2} are of low resolution.
Our audio encoder receives a mel-spectrogram of size $ 16 \times 80 $ derived from $ 16 $ kHz audio with a window size of $ 800 $ and a hop size of $ 200 $.
We employ the Adam optimizer with $ (\beta_1, \beta_2) = (0.5, 0.999) $ and set the learning rate to $ 1 \times 10^{-4} $ for all models.
Training our AVSyncNet is done similarly to SyncNet \cite{prajwal2020lip} on the LRS2 dataset, and then we 
freeze the audio encoder of AVSyncNet and use it in the training of $G_S$ and $G_L$.
At the end of talking face generation, we apply a post-processing step by using VQFR~\cite{gu2022vqfr} to enhance the visual quality and the resolution, aiming to achieve HR videos. 
We train and test our models with a single NVIDIA RTX A6000 GPU.

\section{Experimental Results}
\label{sec:experiments}

\textbf{Dataset.}
We trained our silent-lip generator and talking face generator using Lip Reading Sentence 2 (LRS2) \cite{LRS2} training set as it is a well-known benchmark with extensive subject diversity.
The evaluation was carried out on the LRS2 test set and extended to the LRW \cite{LRW} test set and HTDF dataset~\cite{zhang2021flow} to demonstrate performance on unseen data.

\textbf{Metrics and Baseline.}
For visual quality, we employ widely used metrics: FID \cite{heusel2017gans}, SSIM \cite{wang2004image}, and PSNR.
We also use inter-frame consistency (IFC), presented as a training objective in \cite{zhou2022responsive}. 
This is achieved by calculating the difference between the distances of the consecutive frames in the generated and GT videos.
To evaluate lip synchronization, we follow the literature and use Landmark Distance (LMD) \cite{chen2019hierarchical} in the mouth region and LSE-C \& LSE-D metrics~\cite{prajwal2020lip} to measure the confidence and distance scores through a pretrained model \cite{chung2017out}.
We choose SOTA methods with publicly available codes and models to compare them fairly under the same conditions, as the implementation of the metrics and face cropping strategy before computing the metrics affect the scores.

\begin{table}[tb]
  \caption{Quantitative results on the test sets of LRS2 and LRW.}
  \label{tab:results}
  \centering
  \resizebox{\textwidth}{!}{\begin{tabular}{@{}l|ccccccc | ccccccc@{}}
    \toprule
     & \multicolumn{7}{c}{\textbf{LRS2}} & \multicolumn{7}{c}{\textbf{LRW}} \\
    \midrule
    Method &  SSIM $\uparrow$ & PSNR $\uparrow$ & FID $\downarrow$ & IFC $\downarrow$ & LMD $\downarrow$ & LSE-C $\uparrow$ & LSE-D $\downarrow$ & SSIM $\uparrow$ & PSNR $\uparrow$ & FID $\downarrow$ & IFC $\downarrow$ & LMD $\downarrow$ & LSE-C $\uparrow$ & LSE-D $\downarrow$ \\
    \hline
    Wav2Lip~\cite{prajwal2020lip} & 0.86 & 26.53 & 7.05 & {0.21} & 2.38 & 7.59 & 6.75 & 0.85 & 25.14 & 6.81 & \colorbox{ouryellow}{0.20} & 2.14 & 7.49 & 6.51 \\
    PC-AVS~\cite{zhou2021pose} & 0.73 & 28.24  & 18.40 & 0.46  & 1.93 & 6.41 & 7.52 & 0.81 & \colorbox{ourgreen}{32.25} & 14.27 & 0.38 & 1.42 & 6.53 & 7.15 \\
    EAMM~\cite{ji2022eamm} & 0.69 & 21.01  & 84.65 & 0.51 & 3.54 & 3.31 & 9.93 & 0.71 & 26.22 & 44.16 & 0.48 & 2.61 & 4.32 & 9.04 \\
    VideoReTalking w/ FR~\cite{cheng2022videoretalking} & 0.84 & 25.58 & 9.28 & 0.22 & 2.61 & 7.49 & 6.82 & 0.87 & 27.11 & \colorbox{ouryellow}{5.30} & 0.23 & 2.39 & 6.59 & 7.12 \\
    DINet~\cite{zhang2023dinet} & 0.78 & 24.35 & {4.26} & 0.25 & 2.30 & 5.37 & 8.37 & 0.88 & 27.50 & 8.17 & 0.22 & 1.96 & 5.24 & 9.09 \\
    TalkLip~\cite{wang2023seeing} & 0.86 & 26.11  & 4.94 & 0.24 & 2.34 & \colorbox{ourgreen}{8.53} & 6.08 & 0.86 & 26.34 & 15.73 & 0.26 & 1.83 & 7.28 & 6.48 \\
    IPLAP~\cite{zhong2023identity} & 0.87 & 29.67 & \colorbox{ouryellow}{4.10} & \colorbox{ouryellow}{0.20} & 2.11 &  6.49 & 7.16 & \colorbox{ouryellow}{0.91} & 30.45 & 8.40 & 0.21 & 1.64 & 5.94 & 7.76 \\
    \hline
    Ours w/o FR & \colorbox{ourgreen}{0.95} & \colorbox{ourgreen}{32.64} & \colorbox{ourgreen}{3.83} & \colorbox{ourgreen}{0.16} & \colorbox{ourgreen}{1.13} & 8.41 & \colorbox{ouryellow}{6.03} & \colorbox{ourgreen}{0.92} & \colorbox{ouryellow}{31.45}  & \colorbox{ourgreen}{4.46} & \colorbox{ourgreen}{0.18} & \colorbox{ourgreen}{1.22} & \colorbox{ouryellow}{7.86} & \colorbox{ouryellow}{6.24}  \\
    Ours w/ FR (VQFR) & \colorbox{ouryellow}{0.90} & \colorbox{ouryellow}{31.80}  & 5.23 & 0.27 & \colorbox{ouryellow}{1.36} & \colorbox{ouryellow}{8.52} & \colorbox{ourgreen}{5.83} & {0.90} & 30.21 & 7.05 & 0.21 & \colorbox{ouryellow}{1.41} & \colorbox{ourgreen}{7.92} & \colorbox{ourgreen}{6.00}  \\
  \bottomrule
  \end{tabular}}
\end{table}

\subsection{Quantitative Results}

In \cref{tab:results}, we present quantitative results on test sets of two benchmark datasets, namely LRS2 and LRW. 
We achieve state-of-the-art results in all visual quality metrics excluding PSNR on LRW, 
which is a less informative metric compared to SSIM and FID. 
On IFC, we similarly outperform all compared methods, indicating that our model generates the most consistent videos in terms of temporal information and stability of the faces.

In lip synchronization evaluation, we obtain the best performance in LMD.
Nevertheless, it is important to highlight that LMD is sensitive to the changes in the image, as it does not disentangle synchronization and visual stability.
For instance, affine transformations impact the LMD score even when the lips are synchronized, and vice versa. 
On the LRW dataset, we achieve SOTA results with more reliable confidence and distance metrics for lip synchronization: LSE-C \& D.
On the LRS2 dataset, TalkLip yields a slightly better score than our model with the LSE-C metric.
Nevertheless, we outperform TalkLip and achieve a SOTA result with the LSE-D metric.
All these results indicate the accuracy of our method in terms of visual quality and lip synchronization. 
Similarly, we achieve SOTA results for most metrics on unseen HDTF~\cite{zhang2021flow}, see \appRef{E.1}.

\subsection{Qualitative Results}

\begin{figure}[tb]
  \centering
  \includegraphics[width=\textwidth]{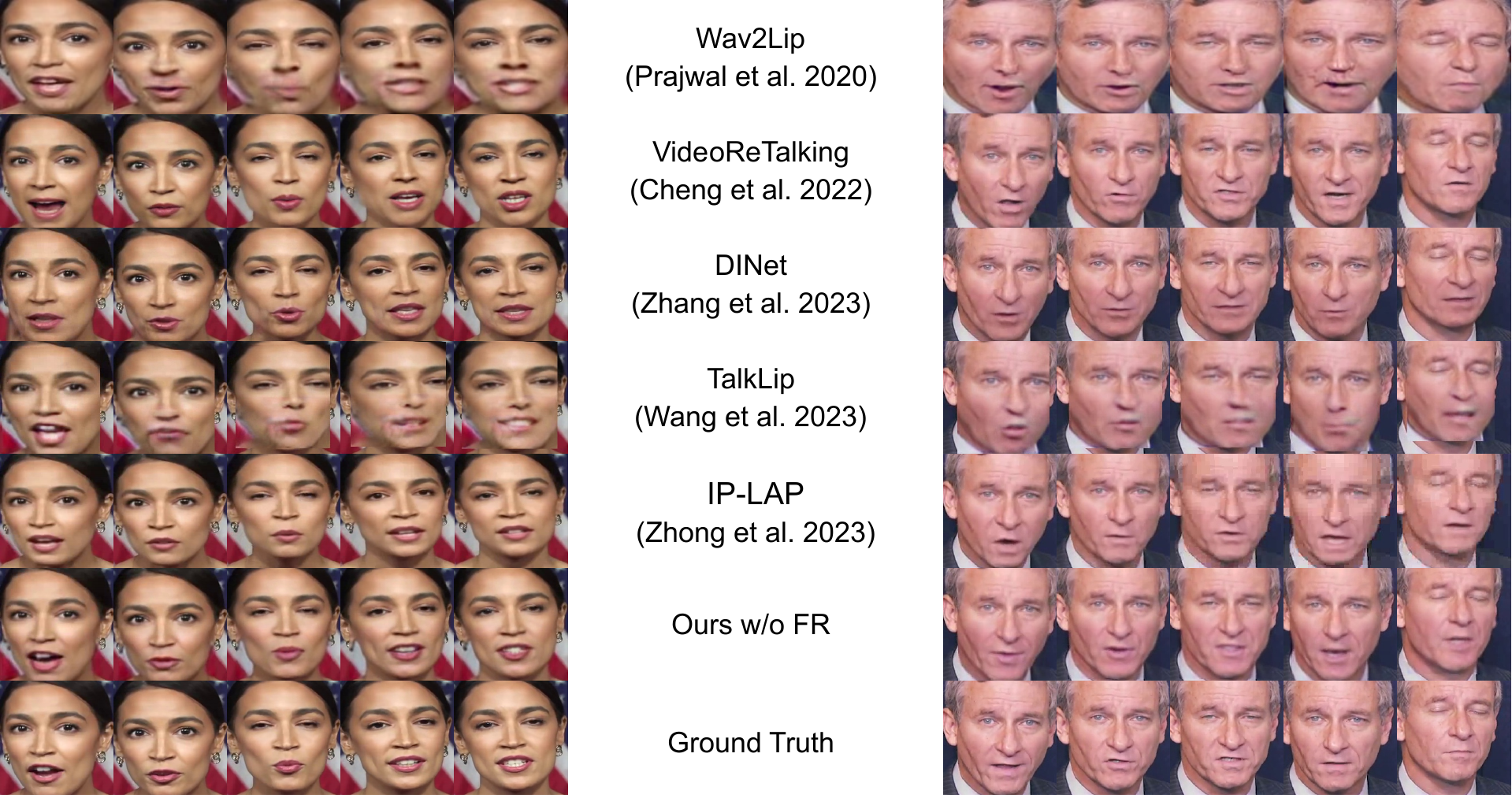}
  \caption{Qualitative comparison with the SOTA methods. Reference videos (from HDTF~\cite{zhang2021flow}) are randomly selected and not seen during training by our model. For more images and videos, please check \appRef{E, F} and {\scriptsize\url{https://yamand16.github.io/TalkingFaceGeneration/}}. 
  }
  \label{fig:qualitative_results}
\end{figure}

In \cref{fig:qualitative_results}, we demonstrate a qualitative comparison with SOTA models and GT data.  
We use their respective publicly available models and generate videos from the HDTF dataset~\cite{zhang2021flow} to compare the models on unseen data since the presented models were trained on the LRS2 dataset, except for DINet (trained with HDTF).
Due to SyncNet and lip-sync loss, TalkLip and Wav2Lip encounter generalization issues, sometimes leading to visual artifacts in the mouth region or face boundaries, especially when the pose of the identity reference differs from the pose reference, despite generating accurate lip movements.
This observation clearly validates the motivation of our contributions.
In contrast, our model generates consistent faces with comparably fewer artifacts, featuring appropriate lip movements that align with both the GT faces and the corresponding audio.
However, VideoReTalking demonstrates comparable lip synchronization and visual quality performance to our model. On the other hand, IP-LAP shows sufficient visual quality, 
while less accurate lip synchronization. 

\begin{table}[tb]

  \caption{Ablation studies on the LRS2 test set. See the text for details.} 
  \label{tab:ablation}
  \centering
  \resizebox{0.8\textwidth}{!}{%
  \begin{tabular}{@{}c|c|l|ccccccc@{}}
    \toprule
Ablation & Setup & Method & PSNR $\uparrow$ & SSIM $\uparrow$ & FID $\downarrow$ & LMD $\downarrow$ & LSE-C $\uparrow$ & LSE-D $\downarrow$ & IFC $\downarrow$\\
\hline
\multirow{7}{*}{Components} & \texttt{A} & $G_L$ + $\mathcal{L}_s$ & 26.349 & 0.853 & 12.25 & 2.408 & 7.116 $\pm$ 1.92 & 7.396 $\pm$ 1.03 & 0.221 \\ 
& \texttt{B} & \texttt{A} + $E_{a, S}$ & 26.614 & 0.868 & 9.82 & 2.325 & 7.271 $\pm$ 1.76 & 7.106 $\pm$ 0.98 & 0.223 \\ 
& \texttt{C} & \texttt{A} + $E_{a, W}$ & 26.590 & 0.869 & 10.56 & 2.278 & 7.220 $\pm$ 1.75 & 7.158 $\pm$0.99 & 0.228 \\ 
& \texttt{D} & \texttt{B} + $G_S$      & 27.180 & 0.872 & 8.16 & 1.741 & 7.752 $\pm$ 1.71 & 6.413 $\pm$ 0.95 & 0.221 \\ 
& \texttt{E} & $G_L$ + $E_{a, S}$ + $G_S$ + $\mathcal{L}_{ss}$ & 31.166 & 0.925 & 5.27 & 1.140 & 8.370 $\pm$ 1.16 & \textbf{6.032} $\pm$ 0.59 & 0.174 \\ 
& \texttt{F} & \texttt{E} + $\mathcal{L}_t$ & 30.658 & 0.917 & 6.24 & 1.250 & 8.260 $\pm$ 1.34 & 6.176 $\pm$ 0.64 & 0.183 \\ 
& \texttt{G} & \texttt{E} + $\mathcal{L}_{at}$ & \textbf{32.755} & {0.949} & {4.02} & {1.135} & {8.382} $\pm$ 1.16 & 6.057 $\pm$ 0.61 & {0.163} \\ 
& \texttt{H} & \texttt{G} w/ AVSyncNet & {32.640} & \textbf{0.952} & \textbf{3.83} & \textbf{1.130} & \textbf{8.410} $\pm$ 0.97 & 6.037 $\pm$ 0.55 & \textbf{0.160} \\ 
\midrule
\multirow{3}{*}{Post-processing} & \texttt{FR1} & \texttt{Setup H} + GPEN & 28.991 & \textbf{0.919} & 58.77 & \textbf{1.197} & 7.625 & 6.457 & \textbf{0.192} \\ 
& \texttt{FR2} & \texttt{Setup H} + GFPGAN & 31.169 & 0.916 & 13.07 & 1.219 & 7.624 & 6.496 & 0.214 \\
& \texttt{FR3} & \texttt{Setup H} + VQFR: \texttt{full model} & \textbf{31.806} & 0.905 & \textbf{5.23} & 1.365 & \textbf{8.528} & \textbf{5.838} & 0.278 \\
\midrule
\multirow{2}{*}{Silent face generation} & \texttt{VRT-S} & VideoReTalking silent data & 22.124 & 0.646 & 33.60 & - & - & - & 0.463 \\
& \texttt{Ours-S} & Our silent data ($G_S$) & \textbf{33.328} & \textbf{0.951} & \textbf{4.41} & - & - & - & \textbf{0.141} \\
  \bottomrule
  \end{tabular}}
\end{table}

\begin{figure}[tb]
  \centering
  \begin{subfigure}{0.40\linewidth}
    \includegraphics[width=\linewidth]{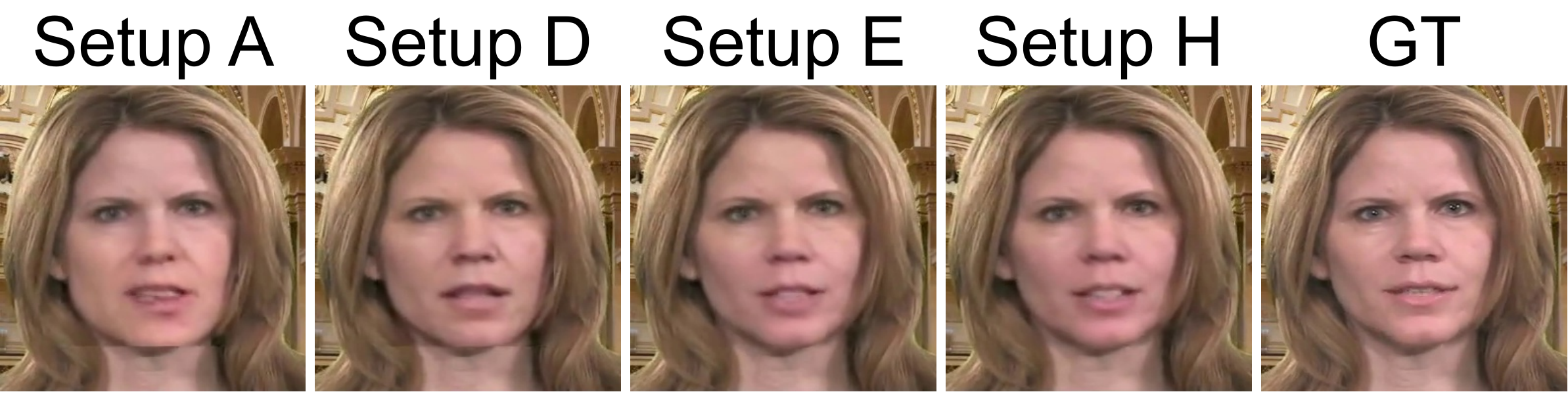}
    \caption{Components.}
    \label{fig:ablation_components}
  \end{subfigure}
  \hfill
  \begin{subfigure}{0.48\linewidth}
    \includegraphics[trim={0 0cm 0 0cm},clip,width=\linewidth]{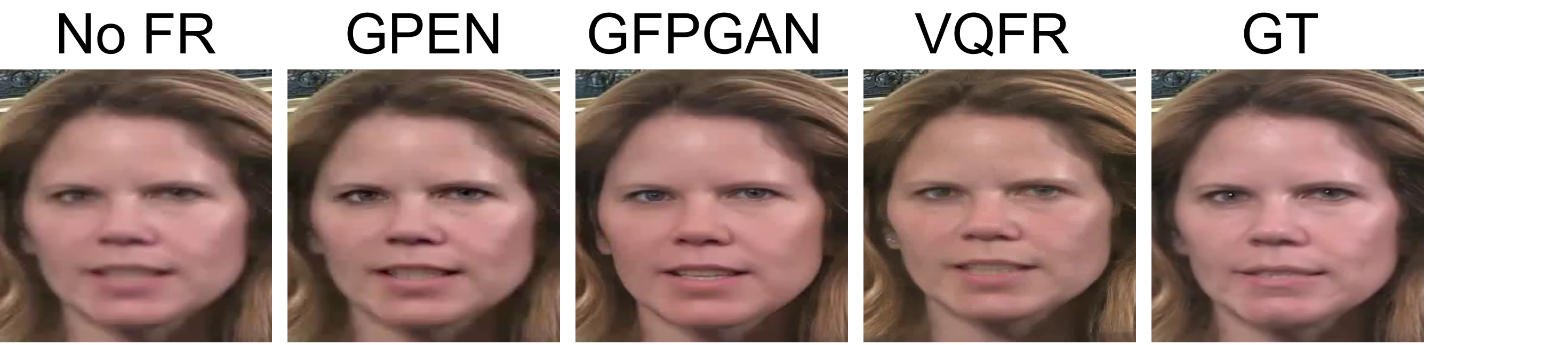}
    \caption{Face restoration methods.}
    \label{fig:ablation_FR}
  \end{subfigure}


  \begin{subfigure}{0.4\linewidth}
    \includegraphics[trim={0cm 0cm 0cm 0cm},clip,width=\linewidth]{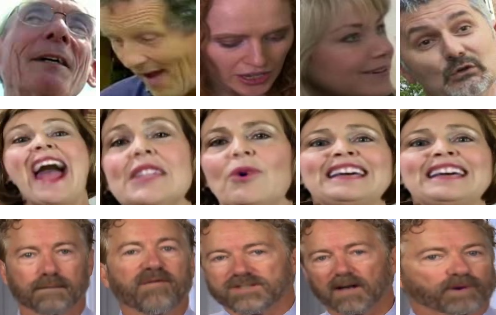}
    \caption{Generated samples for challenging cases: Pose, teeth, beard.}
    \label{fig:challenging_poses}
  \end{subfigure}
  \hfill
  \begin{subfigure}{0.48\linewidth}
    \includegraphics[trim={0 0cm 0 0cm},clip,width=\linewidth]{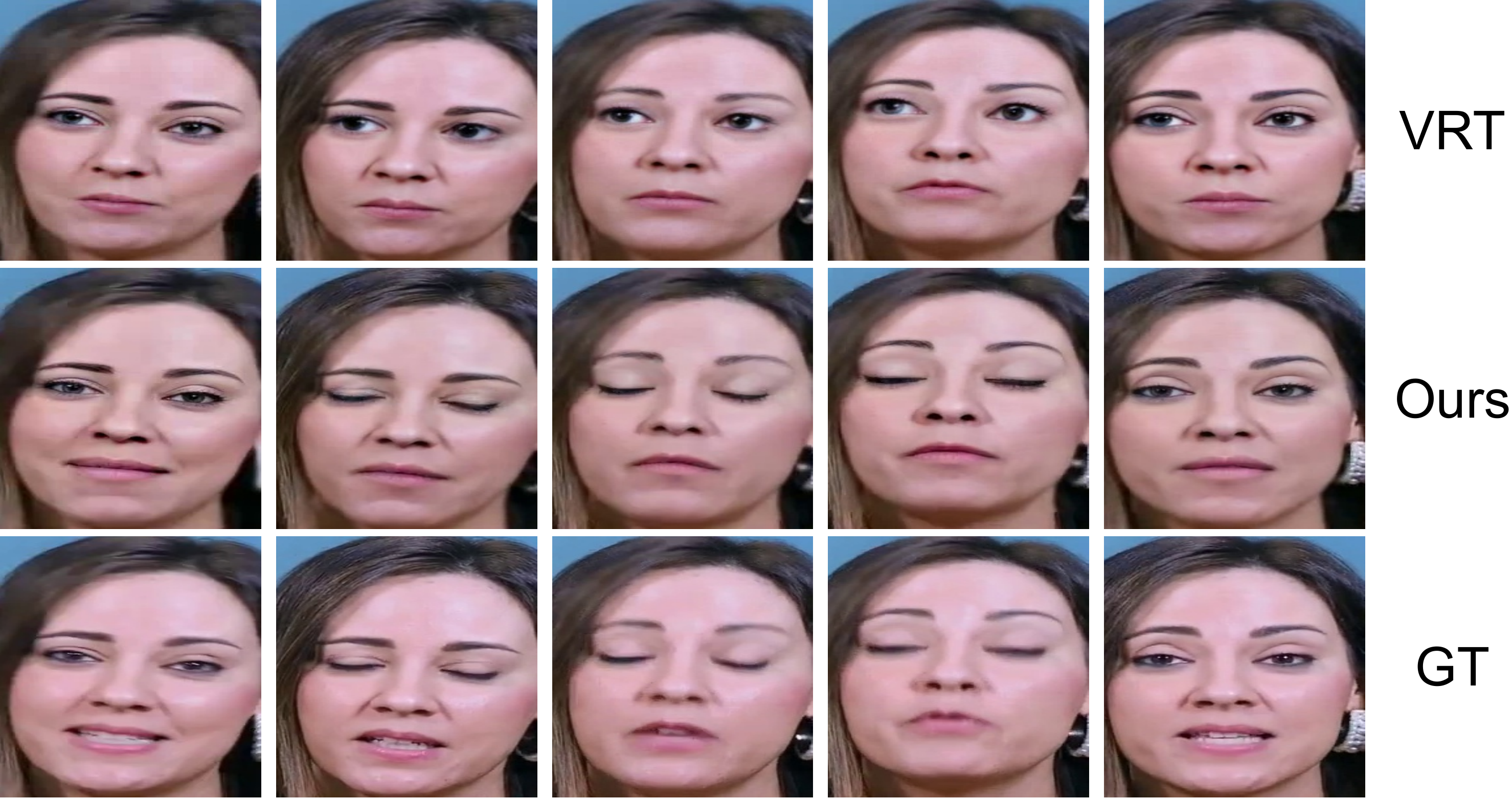}
    \caption{Silent face comparison with VRT~\cite{cheng2022videoretalking}. © European Central Bank (CC BY).}
    \label{fig:silent_faces}
  \end{subfigure}
  \caption{Ablation studies of components (a), face restoration methods (b), and silent face generation (d). (c) demonstrates generated images in challenging cases.}
  \label{fig:ablation_fig}
\end{figure}

\subsection{Ablation Study}

\cref{tab:ablation} and \cref{fig:ablation_components} show a comprehensive ablation study on the LRS2 test set, analyzing the individual impact of our contributions.
We first train our $G_L$ model using SyncNet~\cite{prajwal2020lip} and lip-sync loss~\cite{prajwal2020lip} as a baseline.
As expected, we encountered several issues with unstable training.
Once our model converged after several random seeds, the results (Setup \texttt{A}) show that lip-sync loss can achieve decent synchronization performance, despite lower visual quality and training stability issues.
In this setup, we train the audio encoder as a part of $G_L$.
Replacing this with our pretrained audio encoder enhances the synchronization and visual quality (Setup \texttt{B}).
For further comparison, we also utilize the audio encoder of the Wav2Vec2~\cite{baevski2020wav2vec}, presented in Setup \texttt{C}.
However, it slightly decreases the scores, which validates our hypothesis about training the audio encoder for synchronization purposes.
Thus, we continue with Setup \texttt{B} and add our silent-lip generator $G_S$ to generate silent identity references (Setup \texttt{D}).
$G_S$ alleviates the lip leaking problem, makes the training more stable, and improves the scores noticeably.
We further replace lip-sync loss with our stabilized synchronization loss, yielding drastically improved synchronization and visual quality scores (Setup \texttt{E}).
Moreover, we observed that Setup E shows almost no instability in the training. 
In Setup \texttt{F}, we train Setup \texttt{E} including vanilla triplet loss and it enhances neither visual quality nor synchronization; in fact, it even causes detrimental effects.
This shows the necessity of modifying the triplet loss and introducing the adaptive triplet loss.
In Setup \texttt{G}, replacing vanilla triplet loss with adaptive triplet loss demonstrates a slight improvement in visual quality, while not having a negative impact on lip synchronization. In Setup \texttt{H}, we switch SyncNet with AVSyncNet and achieve slightly better visual quality and lip synchronization.

We compare $G_S$ with VRT~\cite{cheng2022videoretalking} silent face generation approach in \cref{tab:ablation}. Our model surpasses VRT quantitatively and qualitatively (see \cref{fig:silent_faces}), preserving the visual details and identity while modifying the lips.

We compare different face restoration methods as post processing and present the results in \cref{tab:ablation} and \cref{fig:ablation_FR}. 
VQFR surpasses other methods in preserving lip synchronization as well as FID and PSNR. 
However, GPEN shows better performance in the remaining metrics (see \appRef{F.2} for details).
In summary, we employ VQFR for post-processing in our full model.

\subsubsection{AVSyncNet.} 
\begin{figure}[tb]
  \centering
  \begin{subfigure}{0.27\linewidth}
    \includegraphics[width=\linewidth]{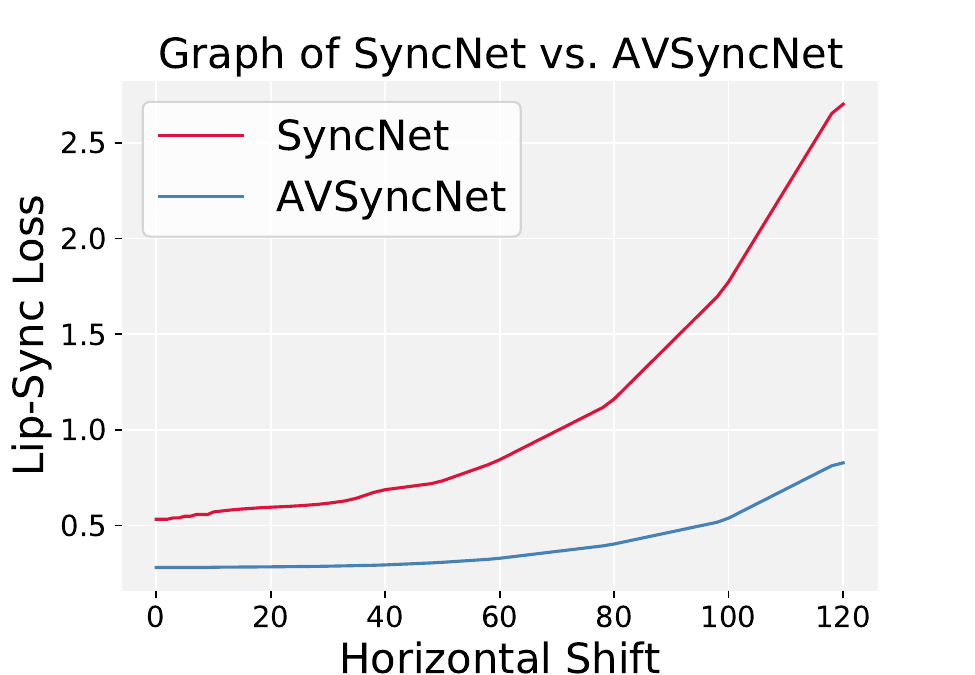}
    \caption{Shifting~\cite{muaz2023sidgan}.}
    \label{fig:ablation_shifting}
  \end{subfigure}
  \begin{subfigure}{0.27\linewidth}
    \includegraphics[trim={0cm 0cm 0cm 0cm},clip,width=\linewidth]{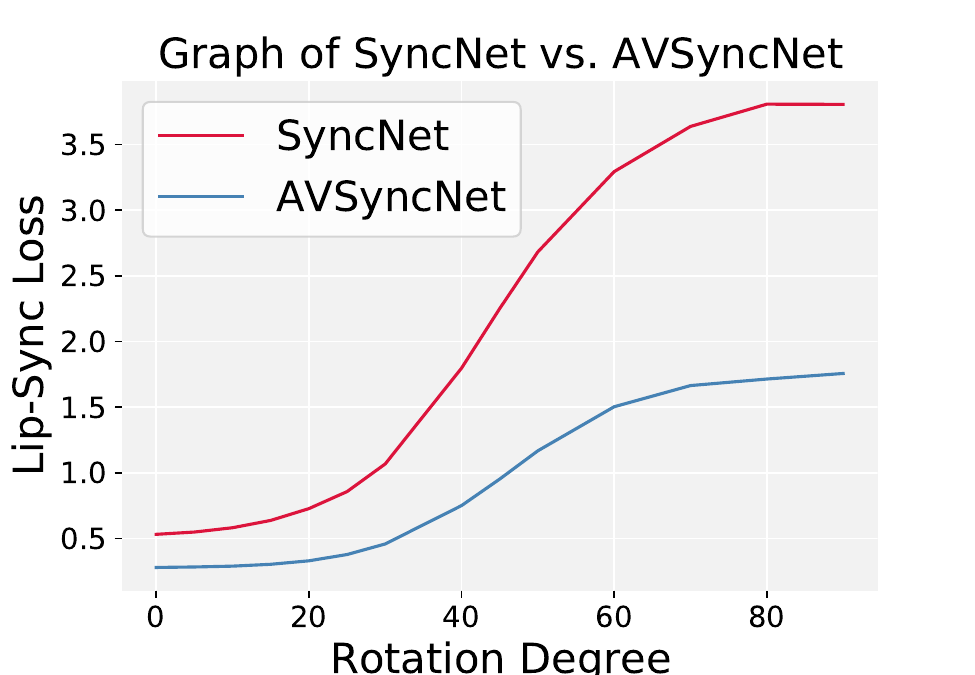}
    \caption{Rotation.}
    \label{fig:ablation_rotation}
  \end{subfigure}
  \begin{subfigure}{0.36\linewidth}
    \includegraphics[trim={0cm 0cm 0cm 0cm},clip,width=\linewidth]{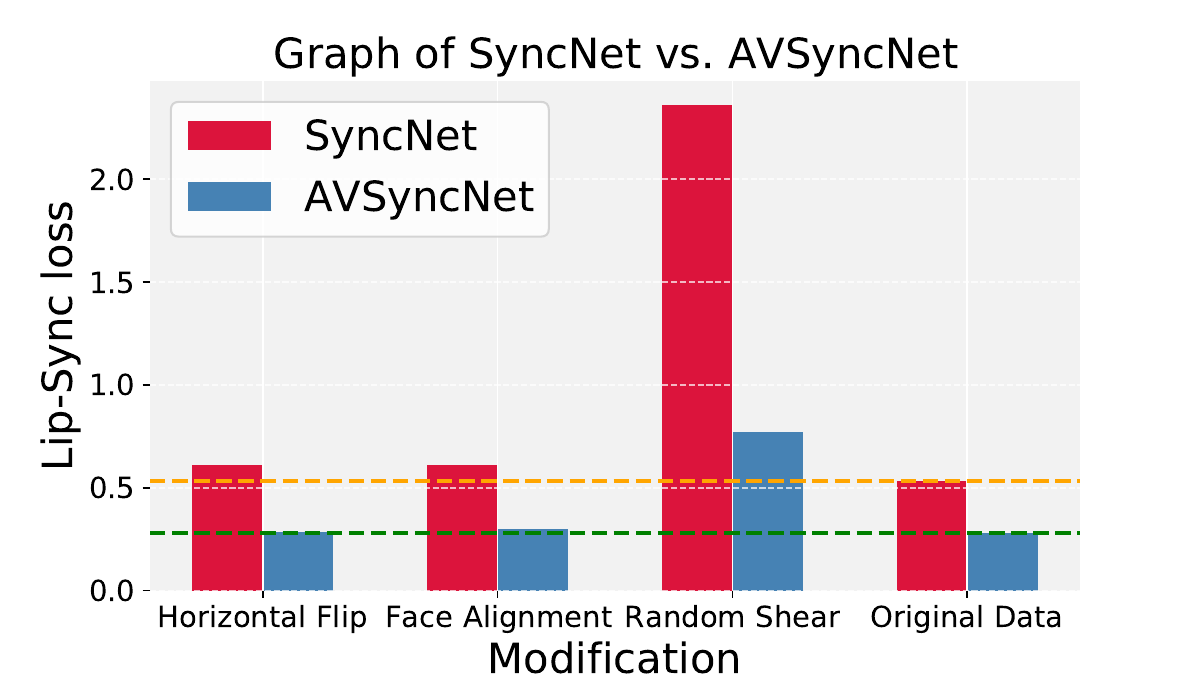}
    \caption{Further analysis.}
    \label{fig:ablation_category}
  \end{subfigure}
  \caption{Graphs show the performance of SyncNet~\cite{prajwal2020lip} and our AVSyncNet on LRS2 GT audio-lip pairs under certain transformations. See \appRef{A} for further details. We apply lip-sync loss~\cite{prajwal2020lip} for the analyses to fairly compare SyncNet and our AVSyncNet independent from our loss function.}
  \label{fig:SyncNet_vs_AVSyncNet}
\end{figure}

Comprehensive experiments on LRS2 GT data pairs highlight AVSyncNet's superior performance (see  \cref{fig:SyncNet_vs_AVSyncNet}). 
Furthermore, it demonstrates strong shift-invariance and robustness against affine transformations in the data due to AVSyncNet's design, particularly emphasizing its effectiveness in focusing on lip synchronization while being less affected by other factors. Moreover, AVSyncNet's performance is not affected by face pose unlike SyncNet~\cite{prajwal2020lip}. \cref{tab:ablation} also demonstrates that our AVSyncNet improves the performance of our talking face generation model and works harmoniously with the proposed $L_{ss}$.

\section{Conclusion}

In this paper, we improve audio-driven talking face generation by identifying problems in current approaches and mitigating them accordingly.
Specifically, we introduce a silent-lip generator to mitigate lip leaking, which is a common problem that harms lip-sync and training stability. 
Furthermore, we propose stabilized synchronization loss along with AVSyncNet, which significantly improves the training stability, lip synchronization performance, and visual quality by solving the problems caused by lip-sync loss and SyncNet.
Experimental results on benchmark datasets and a comprehensive ablation study show the merit of our method and contributions.
Moreover, our detailed analyses reveal the main issues, support our claims, and validate proposed contributions.

\subsubsection{Limitations.}
Despite the notable improvements, SyncNet's and AVSyncNet's unstable nature should be investigated further.
Moreover, face restoration sometimes causes inconsistencies in the video. Silent-lip generator makes teeth invisible in identity references, occasionally resulting in suboptimal teeth generation.

\subsubsection{Ethics \& Social Impact.}
We believe that generating lip-synchronized faces holds significant benefits across a broad spectrum of applications.
However, we acknowledge its vulnerability to potential misuse, particularly deepfake generation.
We will utilize Watermarking 
and prevent uncontrolled usage of our model.

\subsubsection{Acknowledgements.} 
This work was supported in part by the European Commission Project Meetween (101135798) under the call HORIZON-CL4-2023-HUMAN-01-03.

\bibliographystyle{splncs04}
\bibliography{main}
\end{document}